\documentclass[conference]{IEEEtran}
\IEEEoverridecommandlockouts
 \usepackage[numbers]{natbib}
\usepackage{amsmath,amssymb,amsfonts}
\usepackage{algorithmic}
\usepackage{graphicx}
\usepackage{textcomp}
\usepackage{xcolor}
\usepackage[flushleft]{threeparttable}
\usepackage{graphicx}
\usepackage{subfigure}
\usepackage{mathtools}
\usepackage{tikz}
\usepackage{float}
\usepackage{booktabs}
\usepackage{multirow}
\usepackage{comment}
\usepackage[inline]{enumitem}
\usepackage{tabularx}

\usepackage{wrapfig}

\graphicspath{ {./figures/} } 
\begin{document}
\usetikzlibrary{math}
\usetikzlibrary{arrows}
\usetikzlibrary{positioning,calc}
\usetikzlibrary{calc}

\newcommand\wip[1]{{\color{red}#1}}

\title{Learning Non-Stationary Time-Series with \\ Dynamic Pattern Extractions}

\newcommand*{\rebuttal}[1]{{\color{black}#1}}

\newcommand*{\rebuttalBlue}[1]{{\color{black}#1}}

\author{\IEEEauthorblockN{
Xipei Wang$^{*}$}
\IEEEauthorblockA{
xwan2822@uni.sydney.edu.au}
\and
\IEEEauthorblockN{
Haoyu Zhang}
\IEEEauthorblockA{
hzha9377@uni.sydney.edu.au}
\and
\IEEEauthorblockN{
Yuanbo Zhang}
\IEEEauthorblockA{
yzha0492@uni.sydney.edu.au}
\and
\IEEEauthorblockN{
Meng Wang}
\IEEEauthorblockA{
mwan4021@uni.sydney.edu.au}
\and
\IEEEauthorblockN{
Jiarui Song}
\IEEEauthorblockA{
json2540@uni.sydney.edu.au}
\and
\IEEEauthorblockN{
Tin Lai}
\IEEEauthorblockA{
tin.lai@sydney.edu.au}
\and
\IEEEauthorblockN{
Matloob Khushi$^{*}$}
\IEEEauthorblockA{
matloob.khushi@sydney.edu.au}
\thanks{
The authors are with The School of Computer Science, The University of Sydney, Australia.
$^*$Corresponding author.
}
}

\maketitle

\begin{abstract}
    The era of information explosion had prompted the accumulation of a tremendous amount of time-series data, including stationary and non-stationary time-series data. State-of-the-art algorithms have achieved a decent performance in dealing with stationary temporal data. However, \rebuttal{traditional algorithms that tackle stationary time-series do not apply to non-stationary series like Forex trading. This paper investigates applicable models that can improve the accuracy of forecasting future trends of non-stationary time-series sequences. In particular, we focus on identifying potential models and investigate the effects of recognizing patterns from historical data. We propose} a combination of \rebuttal{the} seq2seq model based on RNN\rebuttal{, along with an attention mechanism and an enriched set features extracted via dynamic time warping and zigzag peak valley indicators.} Customized loss functions and evaluating metrics have been designed to focus more on the predicting sequence's peaks and valley points. 
    \rebuttal{Our results show that our model can predict 4-hour future trends with high accuracy in the Forex dataset, which is crucial} in realistic scenarios to assist foreign exchange trading decision making. 
    \rebuttal{We further provide evaluations of the effects of various} loss functions, evaluation metrics, model variants, and components 
    \rebuttal{on model performance.}
\\
\indent\textit{Impact statement}---We investigate pattern-matching techniques in non-stationary time-series data, which closely resemble sensory and financial data \rebuttal{in} our society. Neural networks learn to predict future trends based on patterns or features that it discovers from historical data; therefore, pattern-matching techniques for non-stationary series allow for a more expressiveness model. We introduce three kinds of feature extraction techniques in this work, as a generic framework that \rebuttal{applies} to any general time-series data. We focus our attention on the Forex financial time series, and experimentally we illustrate the best practical model architecture for financial data. We further investigate the impact of loss functions that are tailored to financial series to accelerate model convergence.
\end{abstract}

\section{Introduction}\label{sec:intro}
Over the past decades, improvements in internet technology have led to the explosion of information and massive data accumulation. Businesses, technologies, transportation, media, studies, and many \rebuttal{other sectors had created a tremendous amount of data and became dependent on data-driven approaches}. 
However, in addition to \rebuttal{extracting information from past data, most businesses also want to forecast and infer future information.
Through the use of historical data, we can analyse and extract valuable knowledge,} which allows us to infer and predict future information. 
\rebuttal{Extraction of information will} be a significant help for us to make further informed decisions based on the gathered data.

Timeline based data, also known as time-series data, records the change of attributes of certain events or objects over a period of time. Time series data can be divided into stationary and non-stationary data. Heartbeat, timestamps, train arrival, and departure can all be regarded as stationary time-series data; \rebuttal{their trends tend to be obvious and apparent.} On the other hand, non-stationary data \rebuttal{refers to} 
data that constantly change without obvious regularity, such as stock, wind speed \rebuttal{and} rainfall\rebuttal{. 
Non-stationary data do not have any obvious trends, and they are} often affected by \rebuttal{numerous} internal or external factors beyond prediction. The major problem studied in this paper is the question of \emph{how to forecast future trends in non-stationary time-series data}. 

\rebuttal{In the time-series literature, there are guidelines to classify whether a time-series belongs to the stationary or non-stationary category.}
Previous works \cite{hyndman2018forecasting,alvarez2010energy} point out that patterns extracted from historical time-series data can often be used to improve the performance of future trend forecasting. \citeauthor{bishop2006pattern} \cite{bishop2006pattern} further defines pattern recognition as the process of the automatic regularity discovery in data by using algorithms. 

In this paper, the proposing model can autonomously extract and recognises existing patterns within the \rebuttal{historical} foreign exchange data. The extracted patterns are then further utilised in a neural network model to forecast future trends. Two main types of features are extracted from the input data: (i) the similarity feature \cite{shen2017novel} that denotes the resemblance of the input data against a pre-defined set of common patterns, and (ii) the zigzag indicators \cite{Qi2020Event} which extracts the peak and valley points from the input trend as an expressive financial indicator. \rebuttal{We focus} on examining the predictability and sensitivity of the proposing model with the additional autonomously discovered features. This paper provides the basis for future research on investigating the effectiveness of autonomous pattern extraction and pattern compositions in the area of non-stationary time-series forecast.

Two major objectives are identified in our investigation. The first objective is to structure methods that can recognise a series of typical patterns that residue in the foreign exchange data and verify their corresponding effectiveness with respect to the improvement towards forecasting future trends. Two general categories of patterns, including similarity features and zigzag peak valley indicators, as explained in detail in section \ref{sec:feat_extr}, will be composed together or used separately. The second objective is to design a forecasting model architecture, which can exploit the input non-stationary time-series data stacked with the enriched features set. 
\rebuttal{Our contributions are as follows: (i) we propose a hybrid seq2seq model based on RNN with attention mechanism and investigates its performance against other state-of-the-art methods;
(ii) we identify enrichment features in the non-stationary series and propose automatic extraction of such features in historical data; and 
(iii) we investigate the effect of custom loss functions that identify the peak and valley points in the time-series trend.
We report the best composition of model components, such as loss functions, specific RNN variants, the best combination of patterns and features that obtained the best performance to forecast foreign exchange time series.}

\section{Literature Review}
    A time-series consists of a sequence of observations of which are recorded at some specific timestamps \cite{das1992time}. Time series can further be classified according to their variability along time: if the mean and variance of a time series are constants over time, it is identified as stationary; otherwise, the time series is said to be non-stationary.
    
    \subsection{Patterns in non-stationary time-series}\label{sec:pattern}
    Most time-series data consists of three main common patterns, including (i) the trend which exists when there is a long-term increase or decrease, (ii) the seasonality, which is a repetitive shape of the data and whose occurrence is dependent on seasonal factors, such as a year, a week, an hour, etc., (iii) and the cyclic pattern which refers to the phenomenon where the data rise and fall without fixed frequencies \cite{hyndman2018forecasting}. For example, an important subcategory of derived attributes in stock or foreign exchange data can be categorised by chart patterns such as W, Flag, Wedge, `Head and Shoulder', and `Cup and Handle', of which these patterns have long been utilised as indicators to assist in making trading decisions. 
   
   There are various ways to extract patterns elaborated in related studies. Spearman’s rank correlation, rule sets and sliding window are used in a real-time hybrid pattern-matching algorithm \cite{zhang2010real}. The sliding window enables pattern matching to be performed based on the latest received sub-sequence of time-series data. Therefore, the proposed algorithm can be applied in real-time application. Based on the sliding window, the rank correlation and rule sets are used to distinguish patterns. 
  In addition, the Gini function can be used to determine each pattern's beginning point and length \cite{zhang2009novel}. Dynamic Time Warping is another method that predicts the future trend of the financial series by comparing the price sequence with the predefined patterns \cite{kim2018pattern}.
  %
  \rebuttal{These results implies that we can often interpret the inherent patterns within historical financial data, which would allow us} extracts more information about the pattern from the original data. If this information is input into the model as additional features, it is possible to achieve better performance.

    \subsection{Time-series forecasting}
    Time-series forecasting is the procedure of predicting the future values of a time-series according to the information extracted from its present and past values. 
    
    \subsubsection{Auto-regressive Moving-Average (ARMA) models}
	ARMA models, first introduced by  \cite{whittle1951hypothesis}, have been proved to be effective in short-term forecasting \cite{huang2003short,torres2005forecast}. An ARMA model, being the combination of an Auto-Regressive (AR) model and a Moving-Average (MA) model, \rebuttal{took} advantage of both of the two models by making use of the historical values and lagged errors.
However, since most models designed for stationary time-series forecasting are based on the assumption of constant means and variance, they cannot effectively capture the empirical features in time-series when time-series become non-stationary, such as financial data \cite{petricua2016limitation}. Thus, more advanced techniques for non-stationary time-series forecasting were introduced.

    \subsubsection{ARMA Variants}
    \rebuttal{An improved model had been developed to extend ARMA’s capability on processing non-stationary data}---AutoRegressive Integrated Moving Average (ARIMA) model, which jointly utilises the ARMA model and differencing method. Differencing can convert non-stationary values sequence to stationary residual sequence, which is the strength of ARMA methods. Previous research has proved the effectiveness of the ARIMA model in forecasting non-stationary time-series such as farm price and oil price \cite{nochai2006arima}.
    Another variant is called Generalized AutoRegressive Conditional Heteroskedasticity (GARCH) model, which has shown to be able to exploit the non-linear and non-stationary properties of time-series data such as traffic load \cite{zhou2005network}.
    Although these methods had been applied in specific domains, there exist critical limitations in applying to forecast over-complicated financial data. For example, the ARIMA model is built upon the assumption of constant variance, which does not conform to the high volatility of most trading data \cite{petricua2016limitation}.
    
    \subsubsection{Recurrent Neural Network}
    Recurrent Neural Networks are suitable for handling temporal sequences due to their nature of transferring data in-between cells among each layer, where the order of the sequence can be memorized \cite{rumelhart1986learning}. Studies have been done to compare the performance of the ARIMA models and RNNs, and it is shown that RNNs are able to provide higher satisfactory performances in comparison to the ARIMA models \cite{ho2002comparative}.
    \rebuttal{Long Short-term Memory (LSTM), first proposed by \cite{hochreiter1997long}, outperforms vanilla RNN which} has the problems of exploding gradient and vanishing gradient \cite{hochreiter1997long,sherstinsky2020fundamentals}.  These authors prove that the exploding gradient problem can cause the weights in the neurons being accumulated to an extremely large value as to overflow, while the vanishing gradient problem results in the weights being varied slowly; both of which will lead to the inability for the model to continue to learn from the input data. LSTM, having a more complicated RNN structure and more parameters than that of vanilla RNN, mitigates the above problems by introducing three gates in each neuron that can decide what information can be dropped and what information can be kept.
     \rebuttal{Gated Recurrent Units (GRU),} introduced by \cite{cho2014properties}, is relatively simpler than LSTM with less complicated inner structure, fewer parameters and less computation. Both GRU and LSTM work well in handling long sequence time-series data and there has yet to be any consensus on which one is better among academics.
     %
    Forecasting future trends based on the past events of time-series data falls into the problem of sequence to sequence prediction, which can be handled by a seq2seq model \cite{sutskever2014sequence}. The seq2seq model contains an encoder to obtain the input sequence and a decoder to output the forecasting sequence. Both the encoder and decoder can adopt the aforementioned recurrent neural networks to handle the temporal sequences.
    
     Although the seq2seq model based on RNNs has already achieved a decent performance as evidential by numerous empirical studies, it remains to exhibit the vanishing gradient problem in handling long sequences. The attention mechanism is necessary to further mitigate this shortcoming. The seq2seq model backed with an attention mechanism outweighs the vanilla seq2seq model which follows the encoder-decoder paradigm in handling the long time-series sequence \cite{siddiqui2018generating,kang2018convolve,yang2016stacked}. The attention is a method that, during the decoding processes, allows the decoder to directly connect to the encoder to \rebuttal{focus on the most relevant features from the input signals \cite{michael2019evaluating,yang2016stacked}.}
     
    
    
	
\section{Forex Dataset}
\subsection{Data Collection}\label{sec:data_colle}
    The non-stationary time-series datasets used in this research are the ever-changing price data of the Forex market over several years. These datasets include attributes that describe the datetime, open, close, high, low price, and volume of each time interval of a certain currency pair. All the datasets are obtained with fxcmpy \cite{fxcm_2020}.

    \subsection{Data Analysis}
    The data obtained from Forex are collected at different time intervals: 5, 15, 30 minutes and 1, 2, 4 hours intervals. \rebuttal{For the practical reason} of proposing a realistic model that can forecast relatively long future trends, e.g. several hours or one day, we had chosen 15-minute-interval data with 4 hours as our forecast length (equivalent to 16 data points) of \rebuttal{future trends}. The reason is to balance the trade-off between predicting \rebuttal{an overly} long sequence---which will over complicates the problem for the model to handle, and overly short sequence---which would under-represent the upcoming trend.

    \rebuttal{The collected data, as specified in Section \ref{sec:data_colle}, consists of date, time, various prices and volume. Within these attributes, the time information is uninformative; the open price, which is the close price of the previous time slot, is redundant information that can be disregarded. The close price, being the final state of the current time point, is chosen as the primary input of the model.}

    The data amount is sufficiently large since the data dates back to the 1990s, which contains many representative features for the model to learn. There are also data of various currency pairs, e.g. USD to JPY and GBP to AUD. In this study, we focus on the currency pair of USD to JPY within the \rebuttal{time frame of 2015 to 2019 is chosen for evaluation purposes,} which contain over 120,000 records in total.

    \subsection{Data Processing}
\begin{figure*}[h]
        \centering
        \includegraphics[width=.9\linewidth]{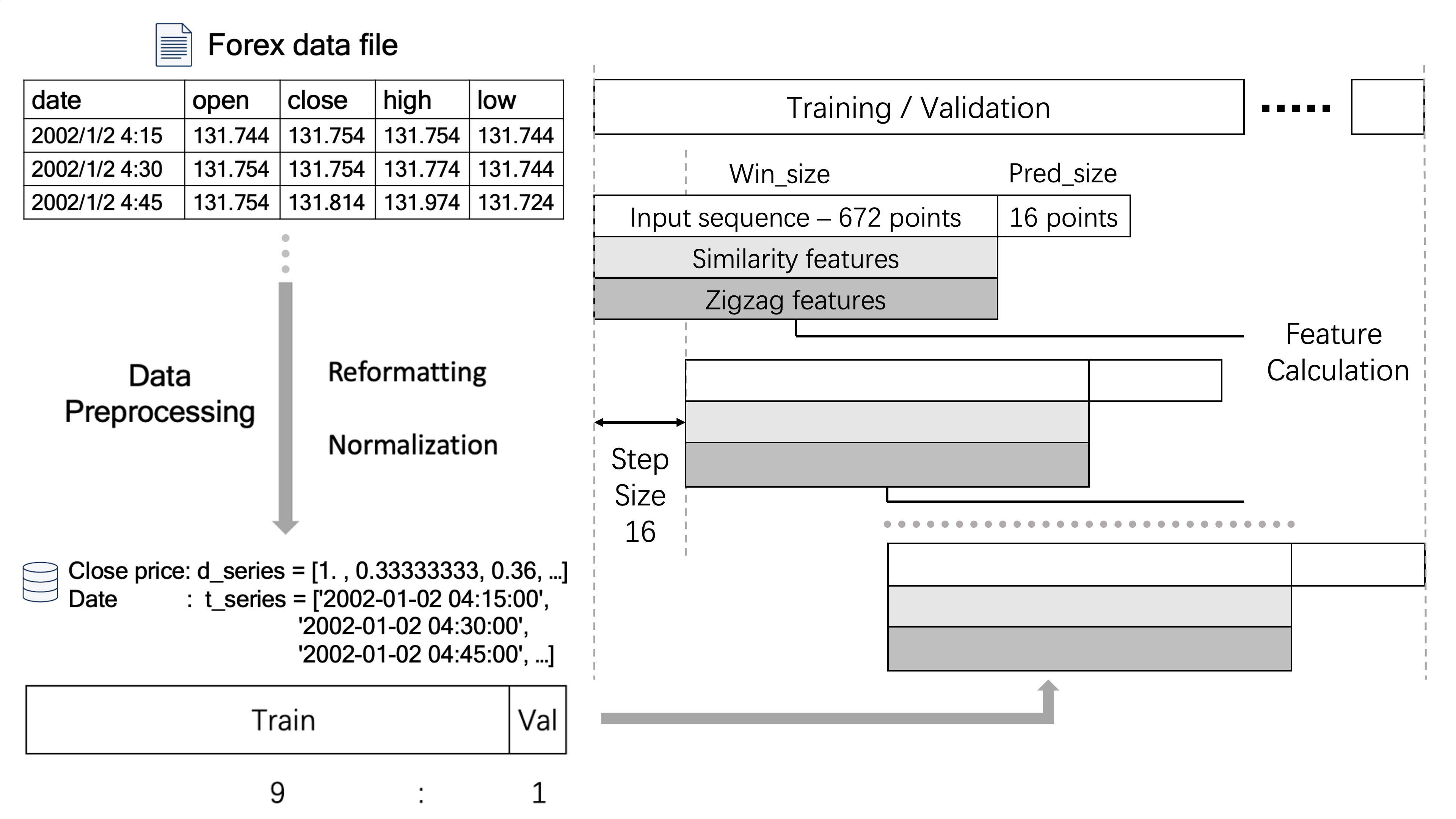}
        \caption{Data pre-processing}
        \label{fig:data_process}
        \end{figure*}

        The \rebuttal{input features are first scaled to a common magnitude such that it enables the model to utilise the learned features across different currency pairs with varying scales. Such an operation enable us to utilise the same underlying patterns across currency-pairs that have different relative scale.} Then, the data are split into a training set and a validation set with a rate of 9:1 for cross-validation. In addition, the data from the training and validation set will be cut by shifting windows into input and output sequence pairs for the seq2seq model. The overall mechanism of data processing is illustrated in Figure  \ref{fig:data_process}. 
\section{Methodology}
        
        \subsection{Model Architecture}
\begin{figure*}[ht]
        \centering
        \includegraphics[width=0.75\linewidth]{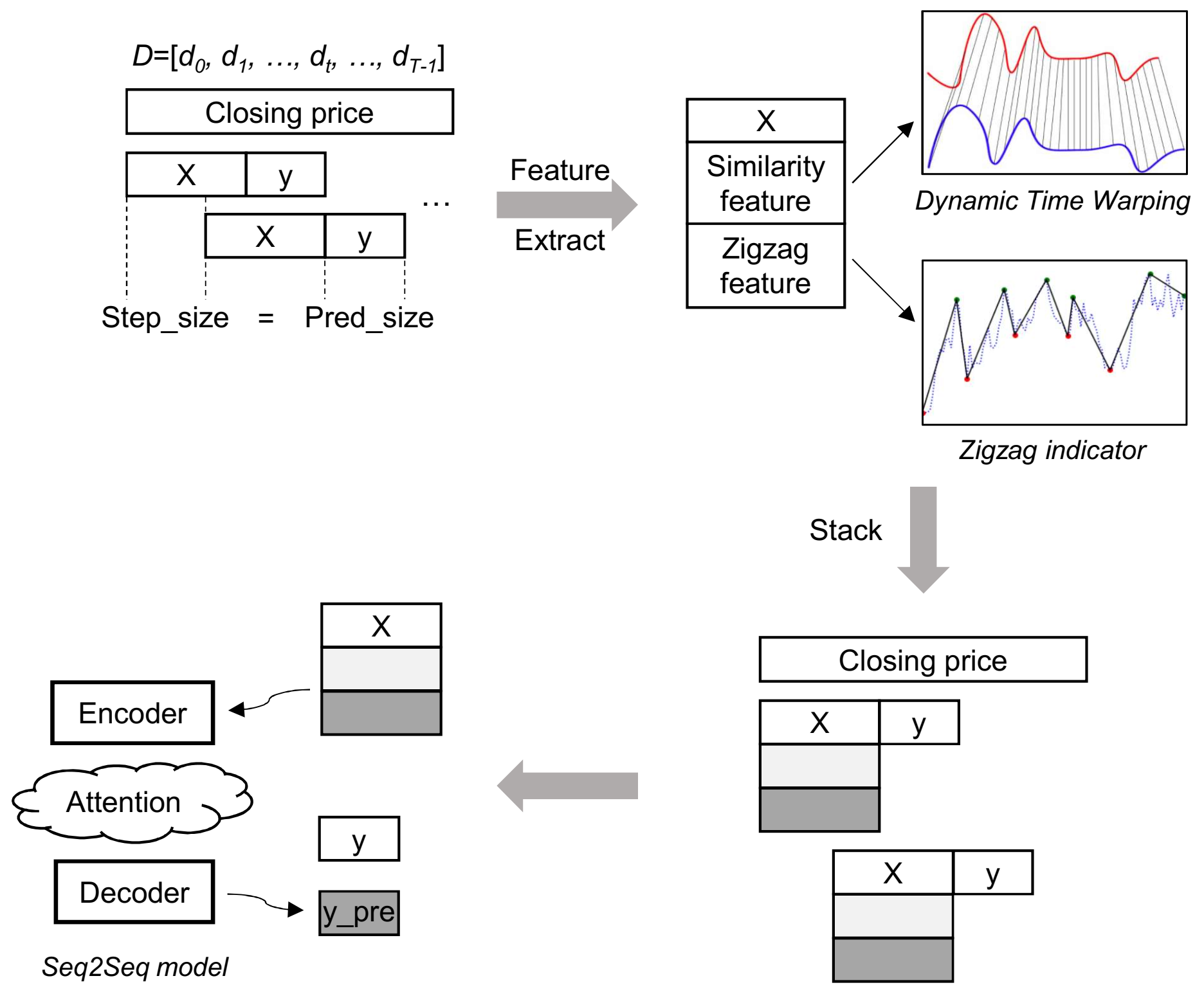}
        \caption{The overall model architecture}
        \label{fig:model_arichitecture}
        \end{figure*}

        The overall model architecture is shown in Figure \ref{fig:model_arichitecture}.
        \rebuttal{Firstly, the series of close price $D=[d_0,d_1,...,d_{T-1}]$ are extracted from the original raw data where $T$ (over 120,000) is the total number of data points.} Each series is sliced as a sliding window with model input $X$, where $|X|=672$, is the input feature in a 7 days period. The output sequence $y$, where $|y|=16$, is the prediction size with a 4 hours period. Each prediction takes a step size of 16. \rebuttal{Then, we begin to extract the similarity pattern and zigzag peak valley features based on each input sequence. Subsequently, these features are stacked onto the input sequence.} Finally, the stacked data and features are input to the seq2seq model, where we predict a sequence from the model and compare it \rebuttal{against the ground truth $y$. We will begin to detail the feature extraction and the structure of the seq2seq model in the following sections.}

        \subsubsection{Feature Extraction}\label{sec:feat_extr}

        \rebuttal{This paper proposes two different approaches in pattern matching and feature extraction for time series data.} The first one is to directly match different segments of the time series data against a predefined set of representative patterns, then subsequently computes similarities score as an additional \rebuttal{feature. Hereafter we will refer to this as \textit{similarity feature}.} Since the proposing model needs to predict the highest and lowest price point in a short horizon to aid trading decisions, the second approach is to recognize the oscillation in time series by marking out the peaks (local maxima) and valley (local minima). \rebuttal{Subsequently, we can use this peak-valley series as an additional feature, which is we will refer to as \textit{zigzag feature}.}
        
        \begin{figure*}[tb]
        \centering
        \includegraphics[width=0.8\linewidth]{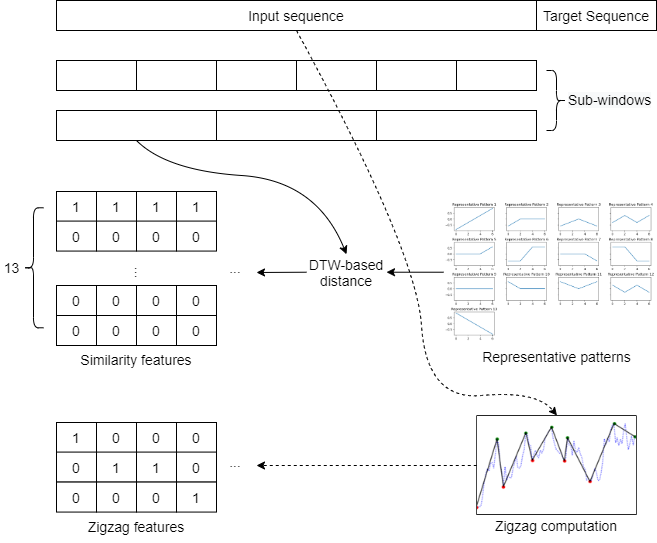}
        \caption{Diagram of feature extraction process}
        \label{fig:feat_extr}
        \end{figure*}
        
\begin{figure}[htb]
         \centering
         \includegraphics[width=0.5\textwidth]{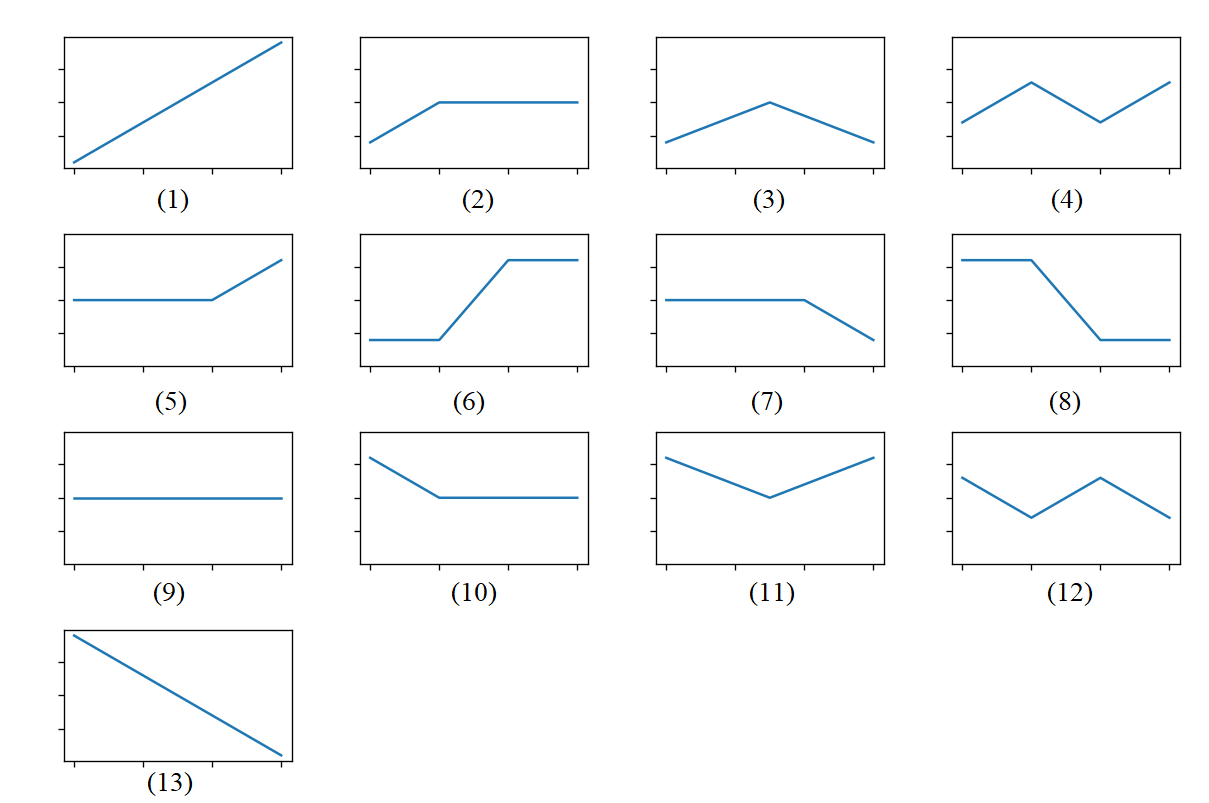}
         \caption{Structure of the 13 representative patterns \cite{kim2018pattern}}
         \label{fig:patterns}
         \end{figure}

        Figure \ref{fig:feat_extr} illustrates the overview of the feature extraction process. For each data pair that contains the input sequence and target sequence, its input sequence will be divided into several sub-windows with equal length. Then, each sub-window is matched against our pre-defined set of representative patterns that are common in the financial market \cite{kim2018pattern}, which produces a distance vector of length 13 that measures the similarities between each sub-window and the patterns (Figure \ref{fig:patterns}). The minimum value in the vector represents the most similar representative pattern, and the vector is further converted to a one-hot encoding form based on the minimum value. The same process is repeated for each pattern to form a similarity feature matrix for the corresponding input sequence. As shown in Figure \ref{fig:patterns}, these representative patterns can be used to represent the basic trend in the trading chart. Thus, the various combination of these representative patterns can effectively represent most of the common chart patterns in Forex or the stock market.

        As shown in Figure \ref{fig:feat_extr}, several groups of sub-windows is used for similarity feature generation, where each group consists of different window sizes. The purpose of this strategy is to extract features with different granularity; thus, features of various lengths can be recognized and extracted. To solve the problem of distance calculation for sub-windows of different lengths, Dynamic Time Warping (DTW) is introduced. DTW is a well-known algorithm for matching two time series of different lengths. In this paper, the Euclidean distance-based DTW algorithm is used to calculate the distance between each sub-window and each representative pattern.
        
\begin{figure}[htb]
         \centering
         \includegraphics[width=0.5\textwidth]{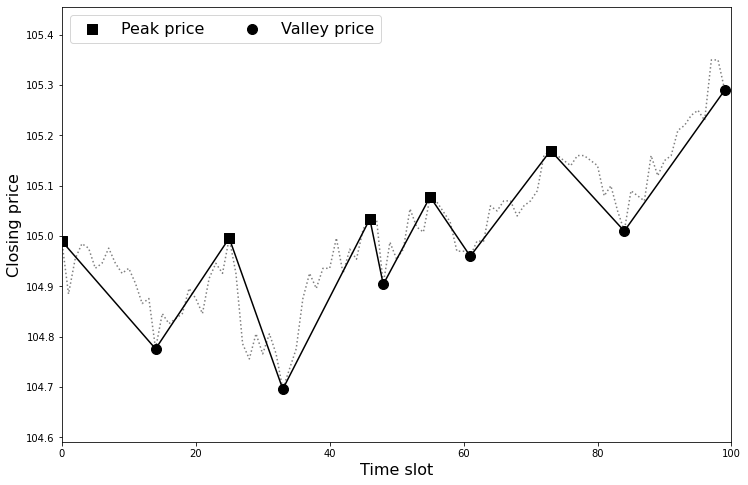}
         \caption{Example of Zigzag indicator}
         \label{fig:zigzag}
        \end{figure}

        The zigzag feature was inspired by the Zigzag indicator commonly used in financial analysis. As shown by an example in Figure \ref{fig:zigzag}, the Zigzag indicator can identify the peak and valley price values in a given price sequence, commonly used by traders for making trading decisions based on the generalised trend. Since the distribution of these peak and valley points can provide important information about future price trend, this paper uses the Zigzag indicator to generate additional features that identify each point in the input sequence as a peak, valley, or other value. Following the same procedure as the similarity feature, the zigzag feature is also one-hot encoded.
          
        As shown in Figure \ref{fig:feat_extr}, in model training or prediction, the similarity features and zigzag features are first computed from the input sequences in each data pair, and then the computed features are stacked with the input sequences to form an input matrix that serves as the input to the model. 

        \subsubsection{Seq2seq Model}
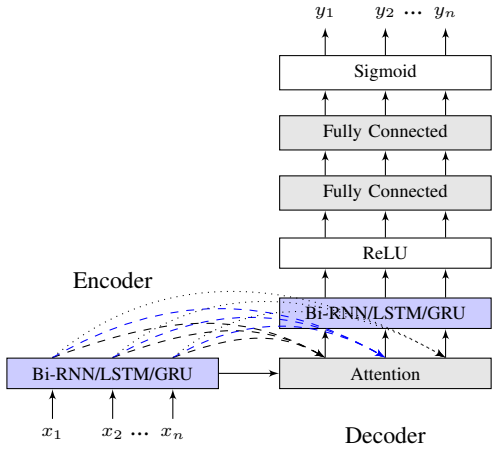
\begin{figure}[h]
                \centering
                    \tikzstyle{rnn} = [draw,fill=blue!20,rectangle, minimum width=8em]
                    \tikzstyle{relu} = [draw,fill=white!20,rectangle, 
                 rectangle, minimum width=8em]
                    \tikzstyle{decoder} = [draw,fill=green!20,rectangle,  rectangle,  minimum width=8em]
                    \tikzstyle{attention} = [draw,fill=gray!20,rectangle,  minimum width=8em]
                    \tikzstyle{dropout} = [draw,fill=gray!20,rectangle, minimum width=8em]
                    \tikzstyle{sigmoid} = [draw,fill=white!20,rectangle, minimum width=8em]
                    \tikzstyle{fc} = [draw,fill=gray!20,rectangle, minimum width=8em]
                    \begin{tikzpicture}[node distance=0.8cm,>=latex']
                        \node[name=Y3] {\scriptsize  $y_n$};
                        \node [left = -0.3 em of Y3] {...};
                        \node[name=Y2, left of =Y3] {\scriptsize  $y_2$};
                        \node[name=Y1, left of =Y2] {\scriptsize  $y_1$};
                        \node[sigmoid, name=sig, below of=Y2] {\scriptsize Sigmoid};
                        \node[fc, name=fc2, below of=sig] {\scriptsize Fully Connected};
                        \node[fc, name=fc1, below of=fc2] {\scriptsize Fully Connected};
                        \node[relu, name=relu1, below of=fc1] {\scriptsize ReLU};
                        \node[rnn, name=decoder, below of=relu1] {\scriptsize Bi-RNN/LSTM/GRU};
                        \node[attention, name=attention1, below of=decoder]{\scriptsize Attention};
                        \node[rnn, name=encoder, left= of attention1.west] {\scriptsize Bi-RNN/LSTM/GRU};
                        \node[name=X2, below of=encoder]{\scriptsize $x_2$};
                        \node[name=X1, left of=X2]{\scriptsize $x_1$};
                        \node[name=X3, right of=X2]{\scriptsize $x_n$};
                        \node [left = -0.3 em of X3] {...};
                        \node[above=of encoder.north]{\small Encoder};
                        \node[below=of attention1.north]{\small Decoder};
                        \draw[->] (X1) -- (encoder.south -| X1);
                        \draw[->] (X2) -- (encoder.south -| X2);
                        \draw[->] (X3) -- (encoder.south -| X3);
                        
                        \draw[->] (encoder) -- (attention1);
                        \draw[->] (attention1) -- (decoder);
                        \draw[->] (attention1.north -| Y1) -- (decoder.south -| Y1);
                        \draw[->] (attention1.north -| Y3) -- (decoder.south -| Y3);
                        \draw[->] (decoder) -- (relu1);
                        \draw[->] (decoder.north -| Y1) -- (relu1.south -| Y1);
                        \draw[->] (decoder.north -| Y3) -- (relu1.south -| Y3);
                        \draw[->] (relu1) -- (fc1);
                        \draw[->] (relu1.north -| Y1) -- (fc1.south -| Y1);
                        \draw[->] (relu1.north -| Y3) -- (fc1.south -| Y3);
                        \draw[->] (fc1) -- (fc2);
                        \draw[->] (fc1.north -| Y1) -- (fc2.south -| Y1);
                        \draw[->] (fc1.north -| Y3) -- (fc2.south -| Y3);
                        \draw[->] (fc2) -- (sig);
                        \draw[->] (fc2.north -| Y1) -- (sig.south -| Y1);
                        \draw[->] (fc2.north -| Y3) -- (sig.south -| Y3);
                        
                        \draw[->] (sig.north -| Y1) -- (Y1);
                        \draw[->] (sig.north -| Y2) -- (Y2);
                        \draw[->] (sig.north -| Y3) -- (Y3);
                        
                        \draw[dashed, ->] (encoder.north -| X1) to [out=20, in = 150] (attention1.north -| Y1);
                        \draw[dashed, ->] (encoder.north -| X2) to [out=20, in = 150] (attention1.north -| Y1);
                        \draw[dashed, ->] (encoder.north -| X3) to [out=20, in = 150] (attention1.north -| Y1);
                        
                        \draw[dashed,blue, ->] (encoder.north -| X1) to [out=30, in = 150] (attention1.north -| Y2);
                        \draw[dashed,blue, ->] (encoder.north -| X2) to [out=30, in = 150] (attention1.north -| Y2);
                        \draw[dashed,blue, ->] (encoder.north -| X3) to [out=30, in = 150] (attention1.north -| Y2);
                        
                        \draw[dotted,black, ->] (encoder.north -| X1) to [out=40, in = 150] (attention1.north -| Y3);
                        \draw[dotted,black, ->] (encoder.north -| X2) to [out=40, in = 150] (attention1.north -| Y3);
                        \draw[dotted,black, ->] (encoder.north -| X3) to [out=40, in = 150] (attention1.north -| Y3);
                        
                        
                    \end{tikzpicture}
                \caption{Seq2seq model structure}
                \label{fig:seq2seq}
            \end{figure}

        The aforementioned input features will be input to the seq2seq model to forecast the incoming trends. Figure \ref{fig:seq2seq} shows the model structure, which contains an encoder and a decoder.
        \rebuttal{The encoder, which takes in the input matrix, is consists of bi-directional recurrent neural network layers. We parameterise this framework into a set of different neural networks, such as RNN, LSTM or GRU, and with different hidden sizes and layers. }
        
       \rebuttal{The decoder, which processes the intermediate encoder output, generates the final output sequence.} The decoder consists of multiple layers, including an attention layer to capture the information from the encoder to enhance the ability to handle the long sequences. \rebuttal{Other layers include multiple parameterized core bi-directional recurrent neural network layers, Rectified Linear Unit (ReLU) layers, fully connected layers and $sigmoid$ activation layers.}
        There is also an optional dropout layer in the encoder and decoder respectively, which is omitted in Figure \ref{fig:seq2seq}, to address the problem of \rebuttal{over-fitting.}
        
        \subsubsection{Loss Functions} \label{sec:loss_func}
	After the model presented in this paper was designed, some initial experiments were performed for predicting future price series of length 96 using an input sequence of length 692. However, \rebuttal{we} found that the training process was relatively slow, and the model's predictions show that it can only predict the general future price direction, but not the precise price fluctuations. In order to solve the problem of slow training, we introduce three new customized loss functions \rebuttal{to force the training process to recognise fluctuations in the target sequence.}
    
    The first customized loss function is called Single Peak Valley (SPV) loss. It adds two penalty terms for the horizontal coordinate difference between peak and valley values in the prediction and target sequences, based on the regular RMSE loss. SPV loss is given by
    \begin{equation}\label{spvrmse}
        \begin{array}{l}
            \mathcal{L}_\textsc{spv} = \mathcal{L}_\textsc{rmse}\times ( 1+\alpha\times pd
            +\beta \times vd),
        \end{array}
    \end{equation}
	where $\mathcal{L}_\textsc{rmse}$ is the typical RMSE loss, $pd$ is the horizontal coordinator difference  between the highest point in the predicted sequence and the target sequence, $vd$ is the horizontal coordinator difference  between the lowest points in the predicted sequence and the target sequence. Both $\alpha$ and $\beta$ are coefficients, which is initially set as 0.5.

	The second customized loss function is called Multi Peak Valley (MPV) loss . It is a variant of $\mathcal{L}_\textsc{spv}$ that adds penalty terms for more than one pair of peak and valley points in the target sequence. MPV loss is given by
	\begin{equation}\label{mpvrmse}
	\begin{array}{l}
    \mathcal{L}_\textsc{mpv}=\mathcal{L}_\textsc{rmse}\times ( 1+
    \sum_{i=1}^k{( \alpha _i\times pd_i  +\beta _i\times vd_i )}),
    \end{array}
    \end{equation}
    where $pd_i$ is the horizontal coordinator difference  between the  $i^{th}$ peak points in the predicted sequence and the target sequence, $vd_i$ is the horizontal coordinator difference  between the $ith$ valley points in the predicted sequence and the target sequence. The peak and valley points of the predicted sequence and the target sequence are selected by using Zigzag indicator with the same parameter setting. By focusing on more peaks and troughs, $\mathcal{L}_\textsc{mpv}$ is expected to provide more accurate feedback to the model, thus further accelerating the training process and helping the model to accurately identify fluctuations in the target sequence.

	The last customized loss function is called Weighted RMSE (WRMSE). It differs from $\mathcal{L}_\textsc{spv}$ and $\mathcal{L}_\textsc{mpv}$ which add penalty terms based on horizontal gap between peak and valley point. WRMSE adds penalty term for each point in the target sequece based on its vertical distance from the average of the target sequence, given by
	\begin{equation}\label{wrmse}
    \mathcal{L}_\textsc{wrmse}=\sqrt{\frac{\sum_{i=1}^n{\left( \hat{y}_i-y_i \right) ^2\left( y_i-\bar{y} \right) ^2}}{n}},
    \end{equation}
where $\hat{y}_i$ is the $ith$ point in predicted sequence and $y_i$ is the $ith$ point in the target sequence,  and $\bar{y}$ is the mean of the target sequence. The purpose of $\mathcal{L}_\textsc{spv}$ and $\mathcal{L}_\textsc{mpv}$ is to more accurately match the horizontal coordinates of peak and valley points, while the purpose of $\mathcal{L}_\textsc{wrmse}$ is to more rapidly match the predicted sequence to the vertical value of each point in the target sequence. \rebuttal{As such, the predicted sequence can be more accurately} matched to the value of the corresponding position in the target sequence.

        \subsection{Evaluation Methods}\label{sec:eva}
        
    \rebuttalBlue{The symmetric mean absolute percentage error (SMAPE) will be used as one of evaluation metric.} Given that this study's objective is to identify market price fluctuations, especially the distribution of peak and valley prices in a short horizon, two \rebuttalBlue{extra} evaluation metrics are designed based on this objective.
    
\rebuttal{    
\subsubsection{Peak Valley RMSE (\textit{PVRMSE})}
      In order to calculate \emph{PVRMSE}, the Zigzag indicator is used to identify the peak and valley points in the predicted and target sequences. }Then, the square errors of corresponding peak and valley points of these two sequences are \rebuttal{calculated according} to their occurrence.  Let $\mathbf{\hat{p}}$ and $\mathbf{p}$ denote the predicted and target peak point sequence, $\mathbf{\hat{v}}$ and $\mathbf{v}$ denote the predicted and target valley point sequence respectively, where $\mathbf{\hat{p}}=(\hat{p}_1,\ldots,\hat{p}_i,\ldots)$ and vice versa.
    Let $M$ and $N$ denote the minimum values among the size of the predicted and target sequence for $\mathbf{p}$ and $\mathbf{v}$ respectively, i.e., $M=min(|\mathbf{\hat{p}}|,|\mathbf{p}|)$ and $N=min(|\mathbf{\hat{v}}|,|\mathbf{v}|)$ where $|\cdot|$ is the cardinality operator.
    Then, \textit{PVRMSE} is given by
    \begin{equation}\label{formula_pvrmse}
    \textit{PVRMSE}=\sqrt{\frac{\sum_{i=1}^M{\left( \hat{p}_i-p_i \right) ^2+\sum_{j=1}^N{\left( \hat{v}_j-v_j \right) ^2}}}{M+N}},
    \end{equation}
    which gives us a measure of the errors in predictions among the peak and valley points.

\subsubsection{\rebuttal{Peak Valley MAE (\textit{PVMAE})}}

    Similarly, another commonly used evaluation metric, Mean Absolute Error (MAE), has been modified for peak valley point matching.
    Peak Valley MAE (\textit{PVMAE}) is given by
    \begin{equation}\label{formula_pvmae}
    \textit{PVMAE}=\frac{\sum_{i=1}^M{\lVert\hat{p}_i-p_i\rVert+\sum_{j=1}^N{\lVert\hat{v}_j-v_j\rVert}}}{M+N},
    \end{equation}
    where the parameters maintain the same mathematical meaning as in \textit{PVRMSE}.

    \subsection{Experiments Setup}
    Three major experiments are carried out in this research, including (i) finding which loss functions are better for accelerating the convergence of the seq2seq model, (ii) investigating which composition of pattern features can help the model obtaining the best performance, and (iii) comparing which RNN variant used in the seq2seq model is the most suitable for the problem of Forex time-series forecasting and whether the attention mechanism is able to improve forecasting performance \rebuttalBlue{and comparing our model with a baseline ANN model and the traditional ARIMA model}. In the following subsections, the common settings shared by \rebuttalBlue{our seq2seq models with different components}, and the respective parameters for each model variant will be introduced in detail.
    \subsubsection{Common Settings}
    The settings about the data, shared by \rebuttalBlue{all the seq2seq model variants}, is described in Table \ref{tab:data_set_setting}.
    \begin{table}[ht]
        \centering
        \caption{Common settings\label{tab:data_set_setting}}
        \begin{threeparttable}
\begin{tabular}{@{}ll@{}}
\toprule
                                                    & Parameters                         \\ \midrule
Data Source$^{[1]}$                                 & USD to JPY                    \\
Time range                                          & 2015-2019$^{[2]}$             \\
Time Interval                                       & 15 minutes$^{[3]}$            \\
Input sequence length                               & 672$^{[4]}$                   \\
Output sequence length                              & 16$^{[5]}$                    \\
Number of epochs                                    & 150                           \\
Learning rate                                       & 1e-4                          \\
\multirow{2}{*}{Number of layers}                   & Encoder: 1                    \\
                                                    & Decoder: 1                    \\
Hidden size                                         & 128                           \\
Batch size                                          & 128                           \\
Dropout rate                                        & 0.0                           \\
Teacher forcing ratio$^{[6]}$                       & 0.0                           \\
Slacked pattern window sizes                        & 672, 336, 96, 48, 24, 12      \\
\multirow{2}{*}{Slacked zigzag differences$^{[7]}$} & 0.0063, 0.007, 0.008, 0.0097, \\
                                                    & 0.012, 0.015, 0.0163, 0.0288  \\ \bottomrule
\end{tabular}
        \begin{tablenotes}
        \item Notes:
        \item $^{[1]}$ Forex data $^{[2]}$ 125000 records $^{[3]}$ 4 points/hour $^{[4]}$ 7days $^{[5]}$ 4 hours
        \item $^{[6]}$ In the decoder, the ratio of feeding the ground truth to the neurons, otherwise the output of previous point
        \item $^{[7]}$ Generate average around 3, 6, ..., 24 peaks and valleys on the  672-length sequence
        \end{tablenotes}        
         \end{threeparttable}
    \end{table}
    
    \subsubsection{Settings for loss function comparison}\begin{wraptable}{r}{0.45\linewidth}
        \caption{Loss function parameters\label{tab:loss_param}}
            \begin{tabular}{@{}cc@{}}
            \toprule
                Loss & Parameters \\ \midrule
                $\mathcal{L}_\textsc{spv}$      & $\alpha=   \beta= 0.5$  \\ \midrule
                \multirow{4}{*}{$\mathcal{L}_\textsc{mpv}$}    & $k=3$ \\
                & $\alpha_1=\beta_1=0.3$  \\
                & $\alpha_2=\beta_2=0.15$ \\
                & $\alpha_3=\beta_3=0.05$ \\ \bottomrule
            \end{tabular}
\end{wraptable}
     To compare the loss functions described in section \ref{sec:loss_func}, the model is trained 150 epochs with the same parameter settings in Table \ref{tab:data_set_setting}. 
    Certain parameters in relation to the certain loss functions are listed in Table \ref{tab:loss_param}.

    
    \subsubsection{Settings for feature comparison}
    
     These model settings refers to our experiments that focus on the generation, stacking of different granularities of pattern features, and other common parameters are illustrated in Table \ref{tab:data_set_setting}. \rebuttal{We study different combinations of features,} including:
        \begin{enumerate*}
            \item close price,
            \item close price and zigzag,
            \item close price and similarity patterns,
            \item close price, zigzag and similarity patterns.
        \end{enumerate*}

\subsubsection{Settings for model component comparison }

    In this experiment, common parameters in Table \ref{tab:data_set_setting} are used. The model is trained 150 epochs for each RNN variant with or without attention layer, the compositions are listed below:
    \begin{enumerate}
        \item Vallina RNN,
        \item Vallina RNN and Attention,
        \item LSTM,
        \item LSTM and Attention,
        \item GRU,
        \item GRU and Attention.
    \end{enumerate}


\section{Results}\label{sec:res}
There are three types of experiments conducted, including the
comparison of (i) loss functions, (ii) feature compositions, and (iii) model components \rebuttalBlue{and baseline ANN and ARIMA}. By comparing the specifically designed loss functions, the most appropriate one can be identified and used to accelerate the whole training process \rebuttal{with improved} performance.

In addition, the comparison of different compositions of pattern features allows us to identify the kind of extracted patterns that are useful for time-series forecasting. Furthermore, \rebuttal{by comparing} core model components (e.g. RNN variants and attention mechanism) the optimised internal structure of the model can be determined for future research.

Additionally, the \emph{PVRMSE} and \emph{PVMAE} evaluation metrics (see Section \ref{sec:eva}), focusing on the most important peak and valley points in the predicting sequence, are adopted among all the experiments to evaluate the seq2seq model.
    \subsection{Loss Function Comparison}\label{sec:res_loss_func}
\begin{figure}[ht]
        \centering
        \includegraphics[width=0.48\textwidth]{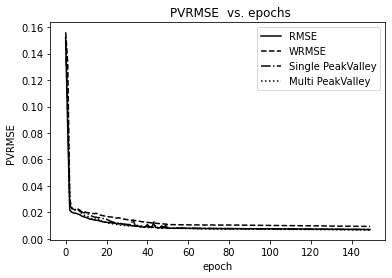}
        \includegraphics[width=0.48\textwidth]{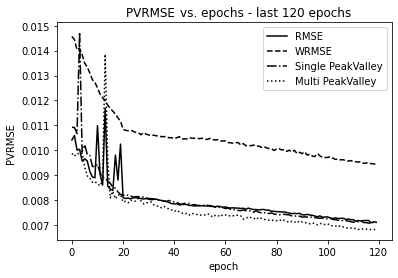}
        \caption{Validation loss comparison among different loss functions (PVRMSE)}
        \label{fig:loss_functions_comparison_PVRMSE}
        \end{figure}

\begin{figure}[ht]
        \centering
        \includegraphics[width=0.48\textwidth]{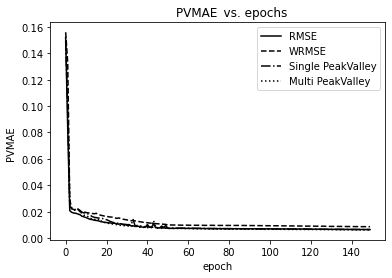}
        \includegraphics[width=0.48\textwidth]{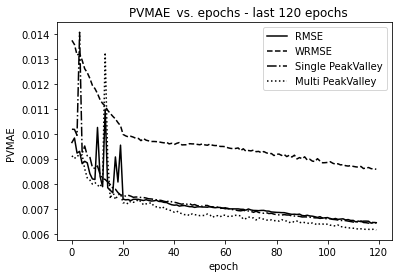}
        \caption{Validation loss comparison among different loss functions (PVMAE)}
        \label{fig:loss_functions_comparison_PVMAE}
        \end{figure}

        This section outlines the loss function that can most effectively helps to accelerate the model training process. We had compared four different loss functions---the $\mathcal{L}_\textsc{rmse}$, $\mathcal{L}_\textsc{wrmse}$, $\mathcal{L}_\textsc{spv}$ and $\mathcal{L}_\textsc{mpv}$. Figure \ref{fig:loss_functions_comparison_PVRMSE}, Figure \ref{fig:loss_functions_comparison_PVMAE} and Table \ref{tab:loss_functions_comparison2}  illustrate that, if trained with the same number of epochs, the model using $\mathcal{L}_\textsc{mpv}$ can provide a prediction with  $\emph{PVRMSE}=4.07 \times 10^{-3}$ and $\emph{PVMAE}=2.60 \times 10^{-3}$ which outweighs the results of using three other loss functions. It means that the $\mathcal{L}_\textsc{mpv}$ is able to speed up the model training by providing more specific information in terms of the peak and valley points in the target sequence. Therefore, this loss function is adopted to be the one that is used in other experiments and in the process of training the final model.
     
        \begin{table}[ht]
        \color{black}
            \centering
            \caption{The comparison of different features\label{tab:loss_functions_comparison2}}
            \begin{tabular}{@{}ccccc@{}}
            \toprule
            Loss functions                       & PVRMSE       & PVMAE  & \rebuttal{SMAPE} \\
            & ($\times 10^{-3}$) & ($\times 10^{-3}$) & \\
            \midrule
            RMSE ($\mathcal{L}_\textsc{rmse}$) & \rebuttal{5.47} & \rebuttal{4.81} & \rebuttal{2.22} \\
            WRMSE ($\mathcal{L}_\textsc{wrmse}$)             & \rebuttal{5.69} & \rebuttal{5.1} & \rebuttal{2.4} \\
            Single PeakValley ($\mathcal{L}_\textsc{spv}$)          & \rebuttal{5.19} & \rebuttal{4.66} & \rebuttal{2.01}\\ 
            Multi PeakValley ($\mathcal{L}_\textsc{mpv}$)          & \rebuttal{\textbf{4.07}} & \rebuttal{\textbf{2.60}} & \rebuttal{\textbf{1.95}}\\
            \bottomrule
            \end{tabular}
        \end{table}

        Additionally, from the result, it can be seen that the performance of the model with $\mathcal{L}_\textsc{spv}$ and $\mathcal{L}_\textsc{mpv}$ is better than that with the other two loss functions. Furthermore, the performance of $\mathcal{L}_\textsc{mpv}$ with the penalty term added for more horizontal coordinates of peak and valley points are better than that of $\mathcal{L}_\textsc{spv}$ with the penalty term added for only one pair of peak and valley points. These findings confirm the assumption of adding suitable penalty terms allow the model to pay more attention to the peak and valley points of the target sequence during the training process. Thus, producing a prediction sequence that is more consistent with the ground truth.
        
        Finally, the result shows that $\mathcal{L}_\textsc{wrmse}$ achieves the worse performance out of all the other three loss functions. \rebuttal{The direct comparison of numeric values might not be fair comparison because the computed loss values might not have the same magnitude as that of the others. Since all hyperparameters provided to the model are the same, it might indicates that the model might require further tuning, such as learning rate, for the $\mathcal{L}_\textsc{wrmse}$ loss function.} Since it is potential to be the one has the ability outweighs the others for the problem of forecasting time-series due to its specific algorithm design (See section \ref{sec:loss_func}), future research can focus on doing more experiments about this loss function by using a different learning rate.
                \subsection{Feature Comparison}\label{sec:fea_com_res}
        The second experiment is to identify which features need to be extracted and stacked to the input sequence of each window. By comparing the model's performance when using combinations of different types of features as model inputs, the experimental results in this section answer the research question of \emph{how to identify and extract the patterns that are implicitly embedded in the Forex trading data}, which allow us to utilise them via enriching the model's input.
        
        As indicated by Table \ref{tab:features_comparison}, \rebuttal{the additional Zigzag features outperform other features across all metrics, whereas the additional DTW similarity pattern features had degraded the performance of our baseline model.
s        In addition, the result also indicates that combining Zigzag and DTW features might requires more training time as the dimensionality of the input sequence are much higher than that of the baseline model.}
        
        \begin{figure}[ht]
        \centering
        \includegraphics[width=0.48\textwidth]{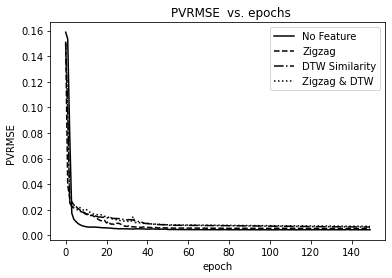}
        \includegraphics[width=0.48\textwidth]{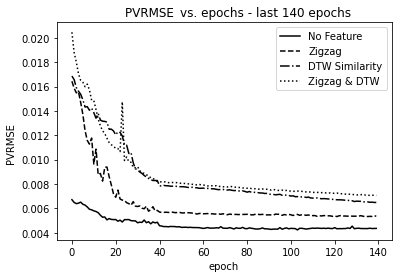}
        \caption{Validation loss comparison among different features (PVRMSE)}
        \label{fig:feature_comparison_PVRMSE}
        \end{figure}
        
        \begin{figure}[ht]
        \centering
        \includegraphics[width=0.48\textwidth]{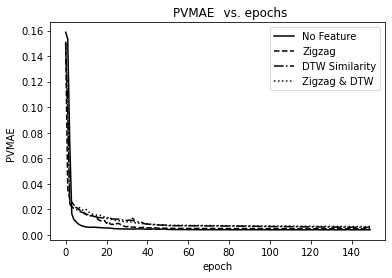}
        \includegraphics[width=0.48\textwidth]{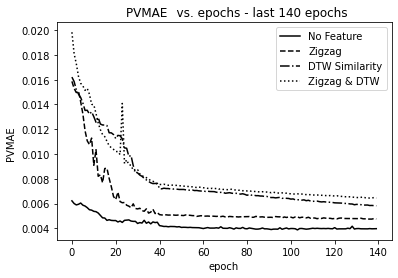}
        \caption{Validation loss comparison among different features (PVMAE)}
        \label{fig:feature_comparison_PVMAE}
        \end{figure}
        \begin{table}[ht]
        \color{black}
            \centering
            \caption{The comparison of different features\label{tab:features_comparison}}
            \begin{tabular}{@{}ccccc@{}}
            \toprule
            Features                       & PVRMSE       & PVMAE  & \rebuttal{SMAPE} \\
             & ($\times 10^{-3}$) & ($\times 10^{-3}$) & \\
            \midrule
            None & \rebuttal{5.19} & \rebuttal{4.66} & \rebuttal{2.01}\\
            Zigzag                         & \rebuttal{\textbf{4.96}} & \rebuttal{\textbf{4.45}} & \rebuttal{\textbf{1.63}}\\
            Similarity Feature             & \rebuttal{5.62}                           & \rebuttal{5.03}         & \rebuttal{1.74}\\
            Zigzag \& Similarity           & \rebuttal{6.29}                           & \rebuttal{5.65}         & \rebuttal{2.33}\\ \bottomrule
            \end{tabular}
        \end{table}
        \subsection{Model Component Comparison}\label{sec:res_model_comp}
        
\begin{figure}[ht]
        \centering
        \includegraphics[width=0.48\textwidth]{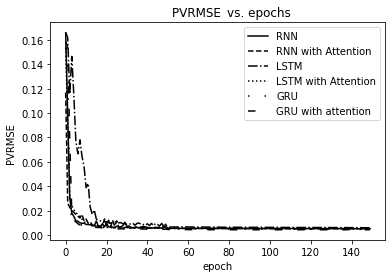}
        \includegraphics[width=0.48\textwidth]{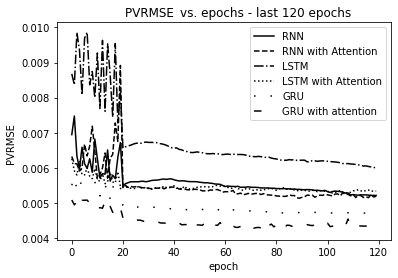}
        \caption{Validation loss comparison among different NN models (PVRMSE)}
        \label{fig:nn_model_comparison_PVRMSE}
        \end{figure}

\begin{figure}[ht]
        \centering
        \includegraphics[width=0.48\textwidth]{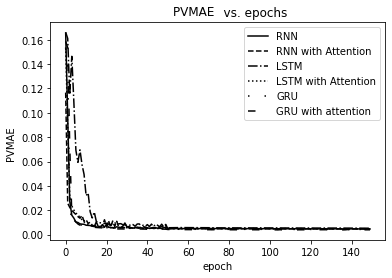}
         \includegraphics[width=0.48\textwidth]{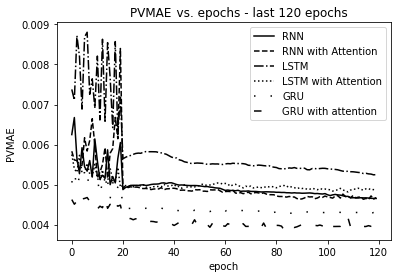}
        \caption{Validation loss comparison among different NN models (PVMAE)}
        \label{fig:nn_model_comparison_PVMAE}
        \end{figure}

        The final experiment \rebuttal{is about varying the core components of the} seq2seq model, which includes 6 different compositions of Vallina RNN, LSTM, GRU and attention mechanism. The result, as shown in Figure \ref{fig:nn_model_comparison_PVRMSE}, Figure \ref{fig:nn_model_comparison_PVMAE} and Table \ref{tab:model_comparison},  \rebuttal{suggests that attention-based Seq2Seq model with GRU has the highest performance (\textit{PVRMSE}: $ 4.96 \times 10^{-3} $, \textit{PVMAE}: $4.45 \times 10^{-3}$) than the other 5 compositions.This is inline with our initial hypothesis since GRU is the more advanced version in the RNN family,} and attention mechanism can improve the neural network dealing with a long sequence. 
        
        \rebuttalBlue{The Diebold Mariano test \cite{diebold1995comparing} has been performed to compare our GRU attention based seq2seq model with a traditional ARIMA, and an ANN model. The ARIMA model consists of the settings of $p=0$, $d=1$, $q=0$, which refer to the auto-regressive, differencing, and moving average terms respectively. The ANN model is composed of 3 dense layers with the dimensionality of (1) input:672, output:128, (2) input:128, output: 32, (3) input:32, output:16. We use Diebold Mariano test function with the settings of MSE differential-loss, $h=|y|^{1/3}+1=16^{1/3}+1=3$, where $h$ is the number of steps ahead of the prediction, and $y$ is the predicting sequence. As shown in Table \ref{tab:dieboldmariano}, the GRU version achieves lower relative error when compared to ANN. The ARIMA model has the highest relative error in all of our paired Diebold Mariano test when compared to the ground truth, with $0.01$ significant level. }
        
        \begin{table}[ht]
        \color{black}
            \centering
            \caption{The comparison of different model components\label{tab:model_comparison}}
            \begin{tabular}{cccccc}
            \toprule
            Model                           & Attention & PVRMSE & PVMAE  & \rebuttal{SMAPE} \\ 
                                            & & ($\times 10^{-3}$) & ($\times 10^{-3}$) & \\
            \midrule
            \multirow{2}{*}{Vallina RNN}    & No & \rebuttal{5.3} & \rebuttal{4.74} & \rebuttal{1.99}\\
                                            & Yes & \rebuttal{5.72} & \rebuttal{5.14} & \rebuttal{1.83}\\ 
            \cmidrule(l){1-5} 
            \multirow{2}{*}{LSTM}           & No & \rebuttal{5.35} & \rebuttal{4.85} & \rebuttal{1.68}\\
                                            & Yes & \rebuttal{5.76} & \rebuttal{5.32} & \rebuttal{\textbf{1.59}}\\ 
            \cmidrule(l){1-5} 
            \multirow{2}{*}{GRU}   & No & \rebuttal{5.35} & \rebuttal{4.85} & \rebuttal{1.63}\\
                                            & \textbf{Yes} & \rebuttal{\textbf{4.96}} & \rebuttal{\textbf{4.45}} & \rebuttal{1.63}\\ 
            \cmidrule(l){1-5}
            \rebuttal{ARIMA}                           & N/A & \rebuttal{492.28} & \rebuttal{492.23} & \rebuttal{199.97} \\
            \cmidrule(l){1-5}
            \rebuttal{ANN}                  & N/A & \rebuttal{12.04} & \rebuttal{11.47} & \rebuttal{2.86}\\
            \bottomrule
            \end{tabular}
        \end{table}

    \begin{table}[ht]
        \color{black}
        \centering
        \newcolumntype{Y}{>{\centering\arraybackslash}X}
        \caption{Diebold-Mariano loss-differential test\label{tab:dieboldmariano}}
        \begin{threeparttable}
        \begin{tabularx}{.9\linewidth}{YYYY}
            \toprule
                Model & GRU+Attention & ARIMA  & ANN \\
            \midrule
                GRU+Attention & - & 123.93$^*$ & 0.53\\
                ARIMA & -123.93$^*$ & - & -120.88$^*$\\
                ANN & -0.53 & 120.88$^*$ & - \\ 
            \bottomrule
        \end{tabularx}
        \begin{tablenotes}
            \item \textbf{Notes}: Negative sign of the statistics implies the second model (row) has bigger forecasting errors.
                $^*$Significant at the $0.01$ level.
        \end{tablenotes}        
         \end{threeparttable}
    \end{table}

    \subsection{Trend Forecasting Result}
    
\begin{figure}[t] 
    \centering
    \includegraphics[width=\linewidth]{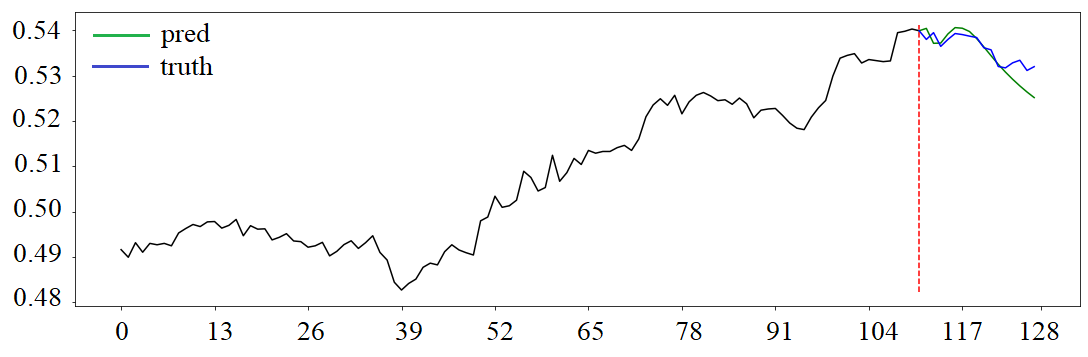}
    \caption{A sample forecasting result of the model}
    \label{fig:Forecasting_model}
    \end{figure}

      Based on the above comparison results, the \rebuttal{combination of $\mathcal{L}_\textsc{mpv}$ as the loss function, attention-based Seq2Seq model with GRU, and the input feature with Zigzag feature is selected as the optimal model.} It is trained for 500 epochs and achieves a performance with the $\textit{PVRMSE}=4.34 \times 10^{-3}$ and $\textit{PVMAE}=3.8 \times 10^{-3}$. Figure \ref{fig:Forecasting_model} shows a sample forecasting result of the optimal model. It indicates that the prediction represented by the green line has a high similarity compared to the ground truth represented by the blue line in both scale and shape, which indicates that this model has the ability in forecasting a relative longer future sequence.
    
\section{Discussion and Conclusion}\label{sec:dis}
    In this research, two types of patterns, including the zigzag peak valley indicator and the time-series change patterns, are recognized, extracted, stacked on the 7-day normalized close price data, and then concatenated as the input for an optimized attention and GRU-based seq2seq model to forecast the price trends over the next 4 hours.
    Several comparative experiments are conducted to select the optimal combination of loss function, feature type, and model structure. In addition, three customized loss functions are designed  to help speed up the model training process. Finally, in order to meet the objective of identifying the peak and valley points in the future price curve, two peak and valley-based evaluation metrics are designed to evaluate model performance. 
    
    Based on the results (refer to section \ref{sec:res}), the two research objectives discussed in section \ref{sec:intro} are achieved. In terms of the pattern, the aim is to verify if adding patterns can improve the forecasting performance or to identify the approaches to add these patterns. \rebuttalBlue{Results in section \ref{sec:fea_com_res} suggest that adding patterns of zigzag indicator has a positive effect for improving the forecast performance.  On the other hand, the negative results from the DTW similarity feature indicates that such a feature set should be utilised in the model with a different approach, e.g. stacking with other indicators, using in different granularities, or stacking multiple layers with varied combination of granularities. }
    
    In addition, practically the peak valley prices are the most important points for financial analysts to make trade decisions. We had persuaded this direction by designing two methods for our investigation. One method is to use \rebuttalBlue{extra} specific evaluation metrics, including PVRMSE, and PVMAE (section \ref{sec:eva}), which focuses on peaks and valleys during the model evaluation. The other one is to use customized loss functions, including $\mathcal{L}_\textsc{spv}$, $\mathcal{L}_\textsc{mpv}$ and $\mathcal{L}_\textsc{wrmse}$ (section \ref{sec:loss_func}), to accelerate the model converging speed by explicitly encouraging the model learning towards the direction of finding peak and valley points. The results in section \ref{sec:res_loss_func} indicate that the loss functions designed in this research do help for speeding up the model learning and improving the model performance.  These two methods are recommended to be used by research with the same purpose.
    
    Finally, the result in section \ref{sec:res_model_comp} illustrates that the combination of GRU with the attention outperforms other combinations of RNN variants and attention being used in our seq2seq model in solving the problem of time-series trend forecasting. \rebuttalBlue{Its performance also exceeds that of the traditional ARIMA, the benchmark ANN model.} Thus, it illustrates that GRU with attention mechanism would be the preferred solution in future studies.
\section{Future Works}
Although there are many achievements in this research \rebuttal{as described in section \ref{sec:dis}, due to the limitations of time and computational resources, we did not investigate every aspects on this line of work. Future works can focus on the following aspects:}

\begin{itemize}
    \item From the financial perspective, find more features that can be extracted to enrich the price \rebuttal{data, and investigate their effects on model performance.}
    \item For the zigzag and similarity pattern features, \rebuttal{perform more experiments on extracting them with different granularities and test their effects on the model with different set of combinations.}
    \item Convolutional Neural Networks  with 1-dimensional convolutional layers can be adopted to examine if they can automatically extract \rebuttal{features, and their effort on model performance.}
    \item Two loss functions, including the SPV loss $\mathcal{L}_\textsc{spv}$ and MPV loss $\mathcal{L}_\textsc{mpv}$, are  parameterized with the allowance of changing the weight coefficients of the error on peak and valley points. More experiments can be carried out to investigate \rebuttal{the set of parameters for the loss functions that would allow us to achieve the highest performance} .
    \item \rebuttal{The prblem with overfitting tends to emerge when we continued to train the model with more epochs.} The model can be modified to add more measures to mitigate this problem. Another measure that can be done is to train the model with more data.
    \item In this research, \rebuttal{we only focus on using a fixed interval for the training and testing data.} Future studies can expand this to the data with different intervals and other currency pairs. Transfer learning can also be carried out to verify if the model can be improved by training different data.
\end{itemize}
\section*{Acknowledgment}

Many thanks to Dr. Matloob Khushi and Dr. Tin Lai for their valuable guidance on technical and financial knowledge. 

\bibliographystyle{plainnat}
\bibliography{references}

\clearpage

\end{document}